\documentclass[letterpaper]{article}

\usepackage{aaai}
\usepackage{times}
\usepackage{helvet}
\usepackage{courier}
\usepackage{url}
\usepackage{graphicx}
\usepackage{subfigure} 
\usepackage{amsmath}
\title{On the Challenges of Physical Implementations of RBMs}
 \author{
     Vincent Dumoulin \and Ian J. Goodfellow \and
     Aaron Courville \and Yoshua Bengio\\
     D\'{e}partement d'informatique et de recherche op\'{e}rationnelle\\
     Universit\'{e} de Montr\'{e}eal\\
     Montr\'{e}al, QC H3C 3J7\\
     \{dumouliv,goodfeli,courvila\}@iro.umontreal.ca\\
     yoshua.bengio@umontreal.ca
 }

\begin{document}
\nocopyright
\maketitle

%------------------------------------------------------------------------------
%                                  ABSTRACT
%------------------------------------------------------------------------------
\begin{abstract}
Restricted Boltzmann machines (RBMs) are powerful machine learning models, but
learning and some kinds of inference in the model require sampling-based
approximations, which, in classical digital computers, are implemented using
expensive MCMC.  Physical computation offers the opportunity to reduce the cost
of sampling by building physical systems whose natural dynamics correspond to
drawing samples from the desired RBM distribution. Such a system avoids the
burn-in and mixing cost of a Markov chain. However, hardware implementations of
this variety usually entail limitations such as low-precision and limited range
of the parameters and restrictions on the size and topology of the RBM.  We
conduct software simulations to determine how harmful each of these restrictions
is. Our simulations are based on the D-Wave Two
computer, but the issues we investigate arise in most forms of physical
computation.
Our findings suggest that designers of new physical computing hardware and
algorithms for physical computers should focus their efforts on overcoming
the limitations imposed by the topology restrictions of currently existing
physical computers.
\end{abstract}

%------------------------------------------------------------------------------
%                                INTRODUCTION
%------------------------------------------------------------------------------
\section{Introduction}
\label{intro}

A restricted Boltzmann machine~\cite{Smolensky86} is a generative model that has
found widespread application~\cite{Hinton06,UTLC+DL+tutorial-2011,Coates2011b}.
Currently, RBMs remain part of the state of the art system for classifying
permutation invariant MNIST~\cite{Hinton-et-al-arxiv2012}. RBMs and other Boltzmann
machines are the dominant means of using deep learning to solve tasks that involve
unsupervised learning and probabilistic modeling, such as filling in missing values
or classification with missing inputs~\cite{Goodfellow-et-al-NIPS2013}.
Unfortunately, the log likelihood of the RBM is intractable~\cite{long10rbm}, and
for other Boltzmann machines most other interesting quantities are intractable as
well. In this paper, we explore the use of quantum hardware to overcome these difficulties.
This approach could possibly unlock the untapped potential of non-restricted Boltzmann
machines.

The model may be trained using sampling-based approximations to the gradient of
the log likelihood~\cite{Younes98onthe,Tieleman08}. However, drawing a fair
sample from the model is also intractable~\cite{long10rbm}.

Drawing samples from an RBM on a classical digital computer is an active area of
research~\cite{Salakhutdinov-ICML2010,Desjardins+al-2010,Cho10IJCNN}. Existing
approaches are based on Markov chain Monte Carlo (MCMC) procedures. The cost of drawing
a fair sample using an MCMC method may be high if the
number of steps required to get a good sample is high. This occurs in practice
because some RBMs represent distributions with modes that are separated by
regions of extremely low probability, which the Markov chain crosses only
rarely.  This is particularly problematic because it interacts with the learning
procedure in a vicious circle: as training progresses, parameters (weights and
biases) gradually become larger, corresponding to sharper probabilities (higher
near training examples, and smaller elsewhere), i.e., corresponding to sharper
modes separated by zones of lower probability.  Since training procedures based
on approximating the log-likelihood gradient require sampling from the model
(usually by MCMC), as training progresses sampling becomes more difficult
(mixing more slowly between modes, i.e., more samples would be required to
achieve the same level of variance in the MCMC estimator of the gradient),
making the gradient less reliable and thus slowing down training.

One possible solution is to construct a physical system whose natural behavior
is to take on states with the desired probability. One may then obtain
the desired samples by observing the behavior of the system, rather than
explicitly performing computations to simulate the dynamics of such a system.
We refer to this approach as ``physical computation''. It is similar in spirit
to ``analog computation'' but we find that term inappropriate in this case,
since the sampled states remain digital. Note that this is different from the
idea of building an RBM ``in hardware''--we are not merely advocating designing
an FPGA that specializes in performing the kinds of digital computations used
for simulating an RBM.

Physical computation is a strategy being actively pursued by D-Wave Systems
Inc.~\footnote{\url{http://www.dwavesys.com/en/products-services.html}}~and DARPA's
UPSIDE program%
\footnote{\url{http://www.darpa.mil/Our_Work/MTO/Programs/Unconventional_Processing_of_Signals_for_Intelligent_Data_Exploitation_(UPSIDE).aspx}}.
In particular, the D-Wave Two system can be viewed as a physical implementation
of an RBM.  Most approaches to physical computation share the property that they
greatly simplify the complexity of a task that is difficult for digital
computers, but also introduce many limitations that digital computers do not
share.  For instance, any physical implementation of an RBM will likely face the issues
of {\em noisy parameters}, {\em limited parameter range} and {\em restricted architecture}.
This paper aims at getting a better understanding of the effect of these
three constraints on the training and performance of the physical RBM and ultimately,
of the feasibility of the physical approach.  In particular, we would like to
address the following questions:

\begin{itemize}
    \item{Which constraint has the worst effect on performance?}
    \item{Under which circumstances can a physical implementation of the RBM be
          reasonably trained?}
    \item{Are there ways to mitigate the degrading effects of constraints
          imposed by physical computation?}
\end{itemize}

Currently, the only practical physical RBM available is the
D-Wave Two system (but see~\cite{Dupret-et-al-1996} for earlier work on
physical computation also associated with Ising models).  It suffers from all
three of the limitations we wish to study. In order to study each limitation in
isolation, we performed a suite of feasibility studies using a {\em simulated}
physical computer, that we implemented in software on a GPU. Using a simulation
allows us to observe what happens when a physical computer has noisy parameters,
but not limited parameter range or architecture restrictions, etc. Because these
experiments are performed in simulation, we do not capture the {\em benefit} of
physical computation: faster, less correlated samples.  Instead, we aim to
characterize the potential {\em detriments} of physical computation. In
particular, by studying each constraint in isolation, we are able to infer their
relative effect on performance and thereby offer guidance for how both hardware
and algorithm designers can best focus their efforts on those properties of
physical computation that impose the greatest barriers to its practical use.

%------------------------------------------------------------------------------
%                                BACKGROUND
%------------------------------------------------------------------------------

\section{Restricted Boltzmann machines}

An RBM is a probabilistic graphical model that
represents a probability distribution over a vector of visible units $\mathbf{v}
\in \{0, 1\}^D$ and a vector of latent variables ({\em ``hidden units''})
$\mathbf{h} \in \{0, 1\}^N$.  As an {\em energy-based model}, the RBM uses an
energy function $E$ to represent a probability mass function:

\begin{equation}
%\begin{split}
    p(\mathbf{v}, \mathbf{h}) = \frac{e^{-E(\mathbf{v},\mathbf{h})}} {Z}, \quad
    Z = \sum_{\mathbf{\tilde{v}},\mathbf{\tilde{h}}}
             e^{-E(\mathbf{\tilde{v}},\mathbf{\tilde{h}})}.
%\end{split}
\end{equation}

In particular, the energy function is
\begin{equation}
    E(\mathbf{v}, \mathbf{h}) = -\mathbf{b}^T\mathbf{v}
                                -\mathbf{c}^T\mathbf{h}
                                -\mathbf{h}^T\mathbf{W}\mathbf{v}.
\end{equation}

This particular form of energy makes the computation of conditional
probabilities trivial:
\begin{equation}
\begin{split}
    p(\mathbf{v} \mid \mathbf{h})
        %&= \frac{1}{1 + \exp(-\mathbf{b} - \mathbf{W}^T\cdot\mathbf{h})} \\
        &= \text{sigmoid}(\mathbf{b} + \mathbf{W}^T\cdot\mathbf{h}), \\
    p(\mathbf{h} \mid \mathbf{v})
        %&= \frac{1}{1 + \exp(-\mathbf{c} - \mathbf{W}\cdot\mathbf{v})} \\
        &= \text{sigmoid}(\mathbf{c} + \mathbf{W}\cdot\mathbf{v}). \\
\end{split}
\end{equation}

Although conditional sampling in an RBM is trivial, sampling from $p(\mathbf{v},
\mathbf{h})$ or from $p(\mathbf{v})$ cannot be done in a single step and
requires the use of Monte Carlo Markov chains, which in general becomes
computationally expensive if the parameters $\mathbf{W}$, $\mathbf{b}$, and
$\mathbf{c}$ are configured in a way that makes the Markov chain mix slowly. 

\section{RBM Learning and Inference}

Given some dataset $\mathbf{V}$ ($\mathbf{v}_{t} \in \mathbf{V}$ for $1\leq t
\leq T$), training an RBM is most commonly done via an approximation to the
gradient of the log-likelihood with respect to the model parameters $\theta_i$,
elements of the parameter vector $\mathbf{\theta}$:
\begin{eqnarray}
\frac{\partial}{\partial\theta_{i}}\left(\sum_{t=1}^{T}\log p(\mathbf{v}_{t})\right)
 = & - & \sum_{t=1}^{T}\left\langle
     \frac{\partial}{\partial\theta_{i}}E(\mathbf{v}_{t},\mathbf{h})\right\rangle
   _{p(\mathbf{h}\mid \mathbf{v}_{t})} \nonumber \\
& + & T \left\langle
     \frac{\partial}{\partial\theta_{i}}E(\mathbf{v},\mathbf{h})\right\rangle
   _{p(\mathbf{v},\mathbf{h})}. \nonumber
\end{eqnarray}
The log-likelihood gradient has two contributions: one in the ``positive phase''
with the expectation over $p(\mathbf{h}\mid \mathbf{v}_{t})$ the model's
conditional hidden unit distribution given the data; the other in the ``negative
phase'' with the expectation over the model's full joint distribution
$p(\mathbf{v},\mathbf{h})$.

While the expectation over the conditional distribution in the clamped condition
is straightforward to compute, the same cannot be said of the expectation over
the joint distribution in the unclamped condition. The evaluation of this
expectation is intractable for all but very small RBMs where the sum over either
all states of the visible layer or all states of the hidden layer is feasible to
compute. In practice, we commonly resort to an approximation to this expectation
via sampling. The persistent contrastive divergence (PCD) algorithm (also known
as stochastic maximum likelihood)~\cite{Younes98onthe,Tieleman08} uses a
persistent Gibbs (MCMC) sampling scheme that sequentially samples from the
conditionals $p(\mathbf{h}\mid \mathbf{v})$ and $p(\mathbf{v} \mid \mathbf{h})$
to recover samples from the joint distribution. These samples are then used in a
Monte Carlo approximation of the negative phase contribution of the log
likelihood gradient. 

While PCD has established itself as probably the most popular method of
maximizing log likelihood in RBMs, it suffers from one important weakness. In
many situations, as learning progresses and the model parameters begin to
increase in magnitude, the Gibbs sampler at the heart of the negative phase
contribution of the gradient can suffer from poor mixing properties. Generally,
it occurs when the hidden and visible activations become highly correlated. Poor
mixing in the Gibbs sampling induced Markov chain leads to poor sample diversity
which in turn leads to poor estimates of the negative phase statistics which
ultimately lead to a poor approximation of the likelihood gradient. This problem
can be somewhat mitigated by increasing sample diversity through the use of
PCD-$k$ (using $k$ Gibbs sampling steps between gradient updates). Other ways
to mitigate the negative phase mixing issue include the use of auxilliary
parameters~\cite{TielemanT2009-small} and tempering
methods~\cite{Salakhutdinov-2010-small,Desjardins+al-2010,Cho10IJCNN}.

The promise of a physical implementation of the RBM is that we entirely
sidestep the difficult mixing problem that
occurs in the negative phase of training by aquiring fair, uncorrelated samples
directly from a physical implementation of the RBM. In the next section we review
the D-wave machine, to our knowledge the only existing physical implementation of an
RBM-like model.

\section{The D-Wave system}

The D-Wave Two system implements an {\em Ising model}~\cite{Ising1925}. Specifically,
it has a signed state vector $s \in \{-1,
1\}^{512}$ and a quadratic energy function \[E(s) = s^T J s + g^T s\] where $J$
is analogous to the weights of a Boltzmann machine and $g$ is analogous to its
biases. The set of Ising model distributions with $\{-1, 1\}$ states is
isomorphic to the set of Boltzmann machine distributions with $\{0, 1\}$ states.
The conversion between the parameters of the two model families is a linear
mapping. An RBM with $\{0, 1\}$ states $h$ and $v$ encoded with weights $W$ and
biases $b$ and $c$ can be converted to use $\{-1, 1\}$ states via the mapping:
\begin{equation}
\begin{split}
    W'   &= \frac{W}{4} \\
    b_i' &= \frac{1}{2} b_i + \frac{1}{4} \sum_j W_{ij} \\
    c_i' &= \frac{1}{2} c_i + \frac{1}{4} \sum_j W_{ji}
\end{split}
\end{equation}
One can draw samples from a Boltzmann machine using the D-Wave Two system
just by performing this linear conversion of the parameters prior to requesting
the sample. The resulting $\{-1, 1\}$ sample may be converted to a $\{0, 1\}$ sample
simply by replacing all instances of -1 with 0.
The choice of parameterization affects the learning dynamics of
stochastic gradient descent, and the Boltzmann
parameterization is usually better, so it is generally best to regard the model
as a Boltzmann machine even if the interface to the sampling hardware uses the
Ising parameterization.

The actual probability distribution sampled by the D-Wave Two system deviates
slightly from $p(s) \propto \exp(-E(s))$.  Moreover, it is difficult to control
the value of $J$ or $g$ precisely. Both effects can be approximated
by adding Gaussian noise to $J$ and $g$. To simulate the D-Wave
Two system with reasonable accuracy, the noise should be added to $J$ once each
time the value of $J$ is changed to a new unique value, but the noise on $g$
should be resampled every time a new sample is drawn\footnote{Andrew Berkley,
D-Wave Principal Scientist, personal communication}.  This is the approach we
take in our GPU-based simulator of the D-Wave Two system.  (One complication we
do not attempt to model is that if the same value of $J$ is requested twice, the
error on $J$ should be the same both times--it is not truly noise, but rather a
deterministic error that has a Gaussian distribution when compared over multiple
points in $J$ space) Other approaches to physical computation, such as those
explored by DARPA's UPSIDE program, face similar issues with noise.

The D-Wave Two system also imposes restrictions on the magnitude of each
individual element of $J$ and $g$.  This is common to most approaches to
physical computation.

\begin{figure}[ht]
    \centering
        \includegraphics[width=\columnwidth]{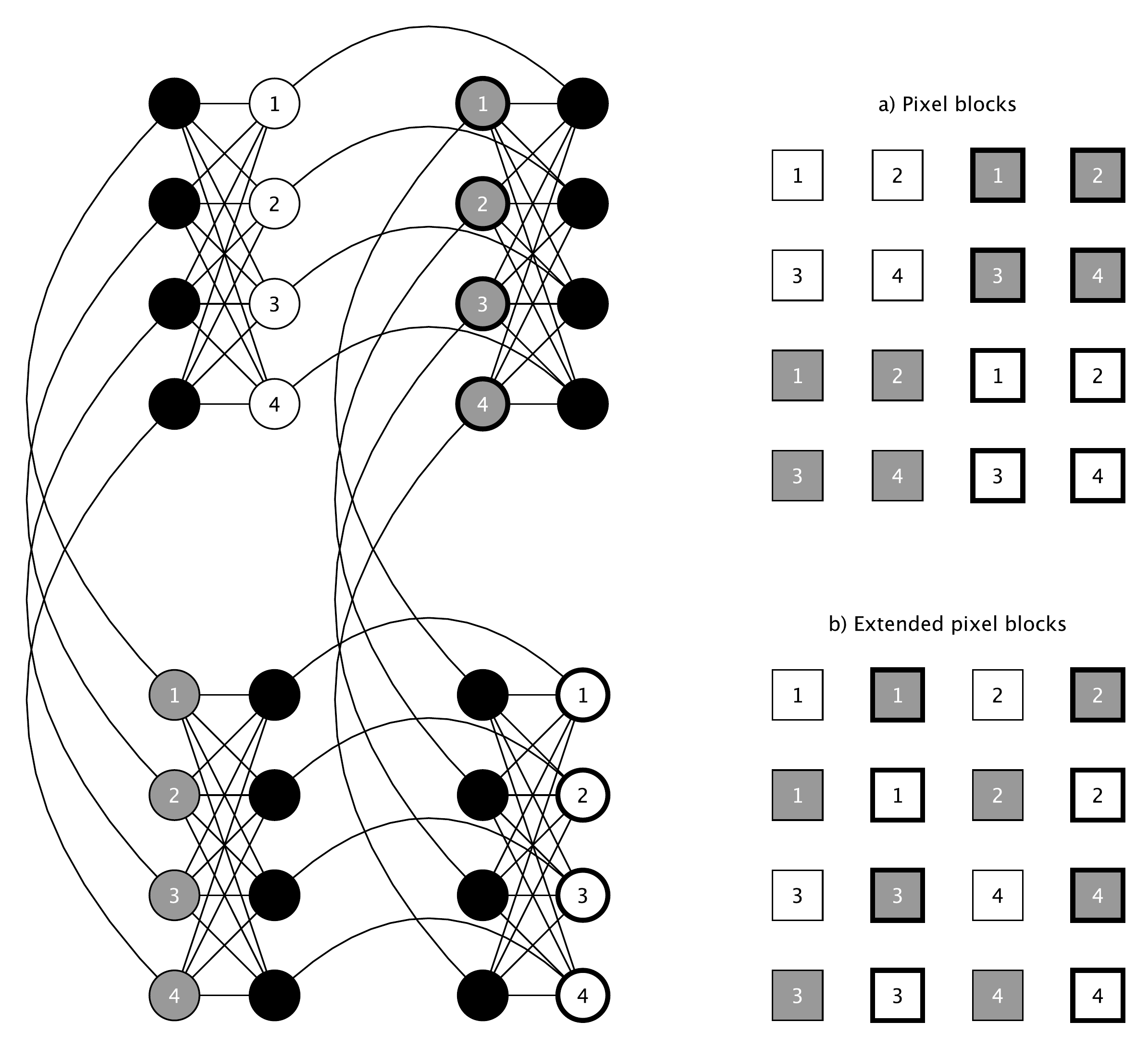}
    \caption{Two different ways of mapping pixels of an image to visible units
             of an RBM with chimera connectivity. \emph{Pixel blocks}
             involves mapping adjacent 2$\times$2 blocks of pixels to
             adjacent, fully-connected groups of units while respecting the
             relative positions of blocks of pixels. \emph{Extended pixel
             blocks} make the pixel blocks
             overlap, to capture more long-range relationships.}
    \label{fig:pixels_to_chimera}
\end{figure}

Finally, many elements of $J$ are constrained to be zero. This is because the
various elements of the state vector are physically laid out in a 2-D grid, and
only nearby elements can interact with each other. Specifically, the
connectivity of the graphical model is constrained to be a chimera graph as
illustrated in Fig.~\ref{fig:pixels_to_chimera}.  We observe that this
chimera graph can be partitioned to form a bipartite graph. Under such a
partition, the D-Wave Two system comes very close to being an RBM. The only
difference between this model and an RBM is that the noise on the biases causes
the biases to be random variables rather than parameters of the model.

% \begin{figure}[ht]
%     \centering
%         \includegraphics[width=\columnwidth]{chimera_2_by_2_unrolled.pdf}
%     \caption{The D-Wave chimera graph seen from two different perspectives.
%              On left is the traditional 2D representation. Its bipartite nature
%              allows us to see it as an RBM with limited connectivity (right part
%              of the figure). Units were colored according to which part of the
%              partition they belong to.}
%     \label{fig:chimera_architecture}
% \end{figure}

\citeauthor{Denil-wkshp-2011} have also explored the use of D-Wave hardware for
training RBMs. Like our work, their work is primarily a feasibility study based
on software simulations. Their approach differs from ours in three respects: 1)
We partition the D-Wave Two system into visible and hidden states using a
partitioning that makes the chimera graph bipartite, so the hardware implements
an RBM. \citeauthor{Denil-wkshp-2011}  used a different partitioning that allowed
visible-visible and hidden-hidden interactions. 2) We train using sampling-based
approximations to the log likelihood gradient, while they train using empirical
derivatives of an autoencoder-like cost function. 3) Our focus is on
understanding how detrimental each of the limitations of the D-Wave hardware is
in isolation, while \citeauthor{Denil-wkshp-2011}  focus on devising an algorithm
that works reasonably well with all limitations in place simultaneously.

\begin{figure}[ht]
    \centering
        \begin{tabular}{cc}
            \includegraphics[width=0.47\columnwidth]{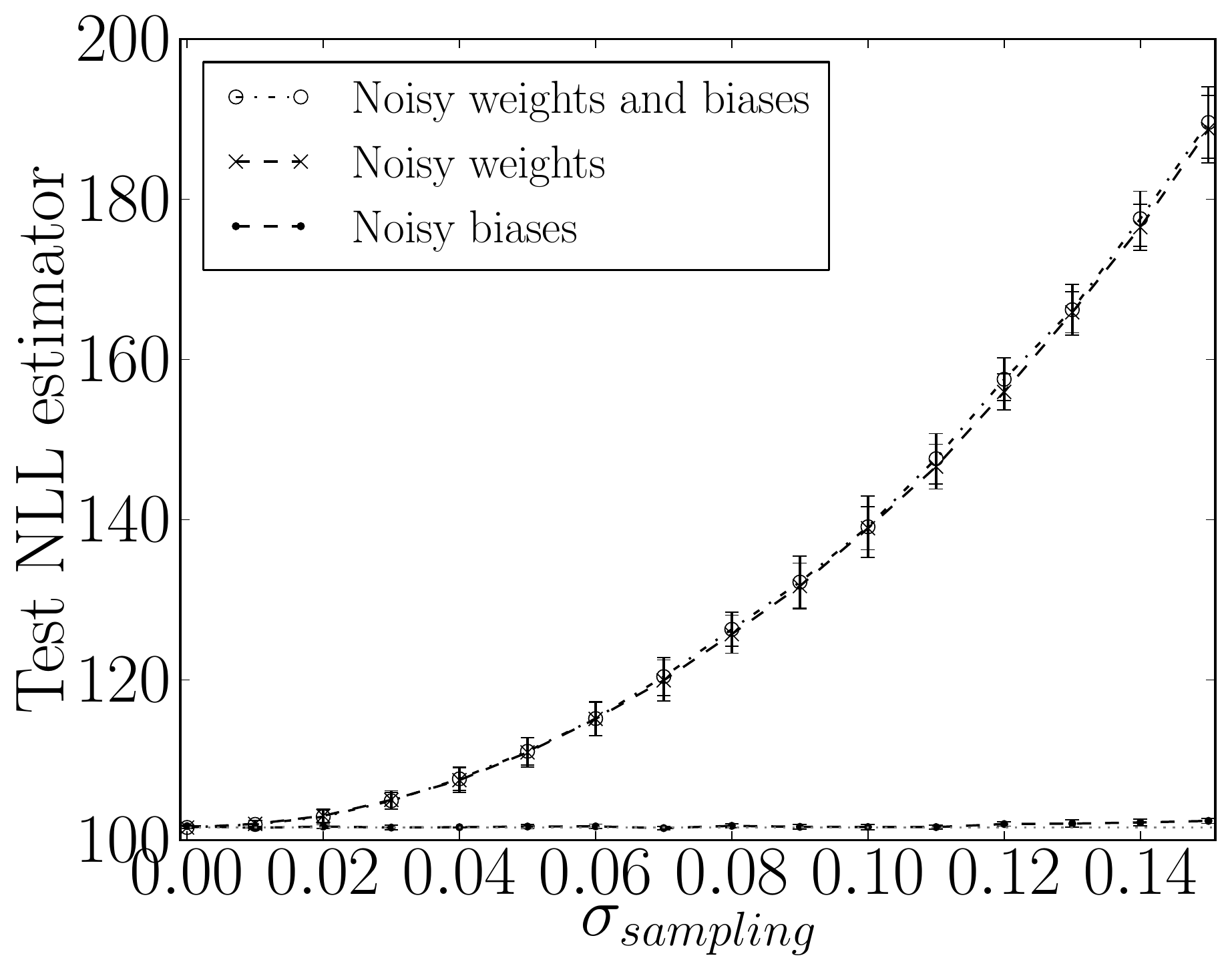} &%
        \includegraphics[width=0.47\columnwidth]{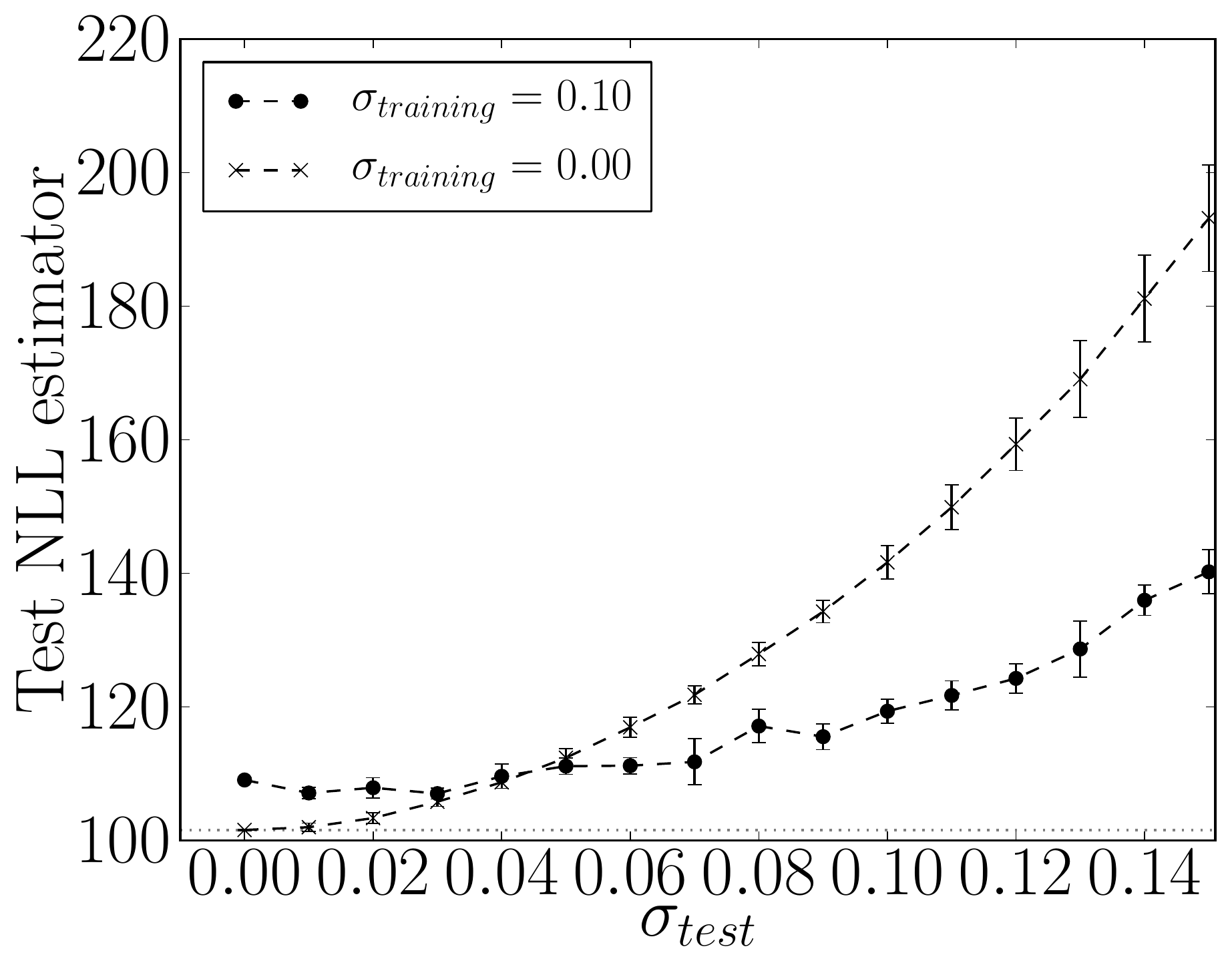}\\
            \includegraphics[width=0.47\columnwidth]{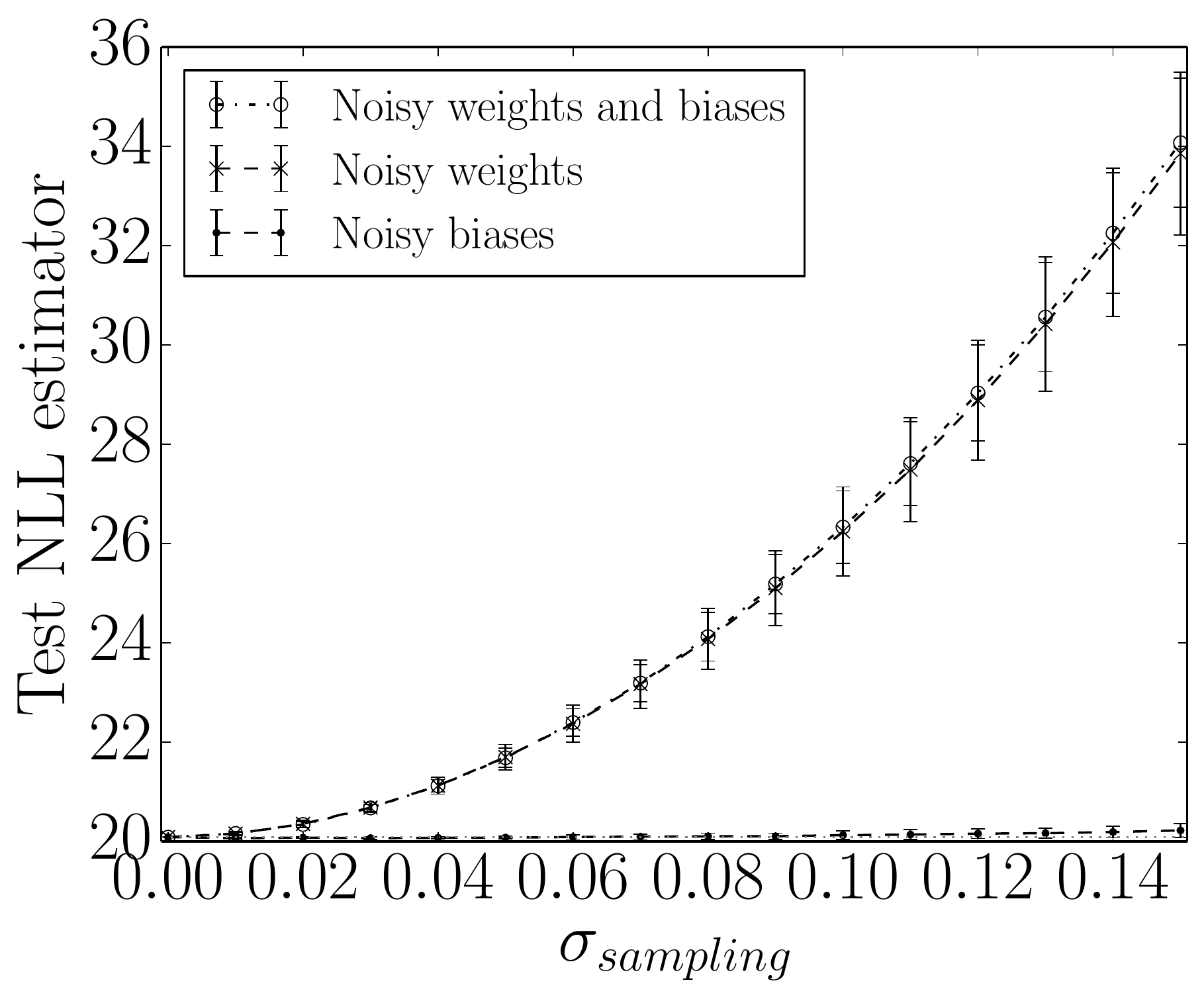} &%
	\includegraphics[width=0.47\columnwidth]{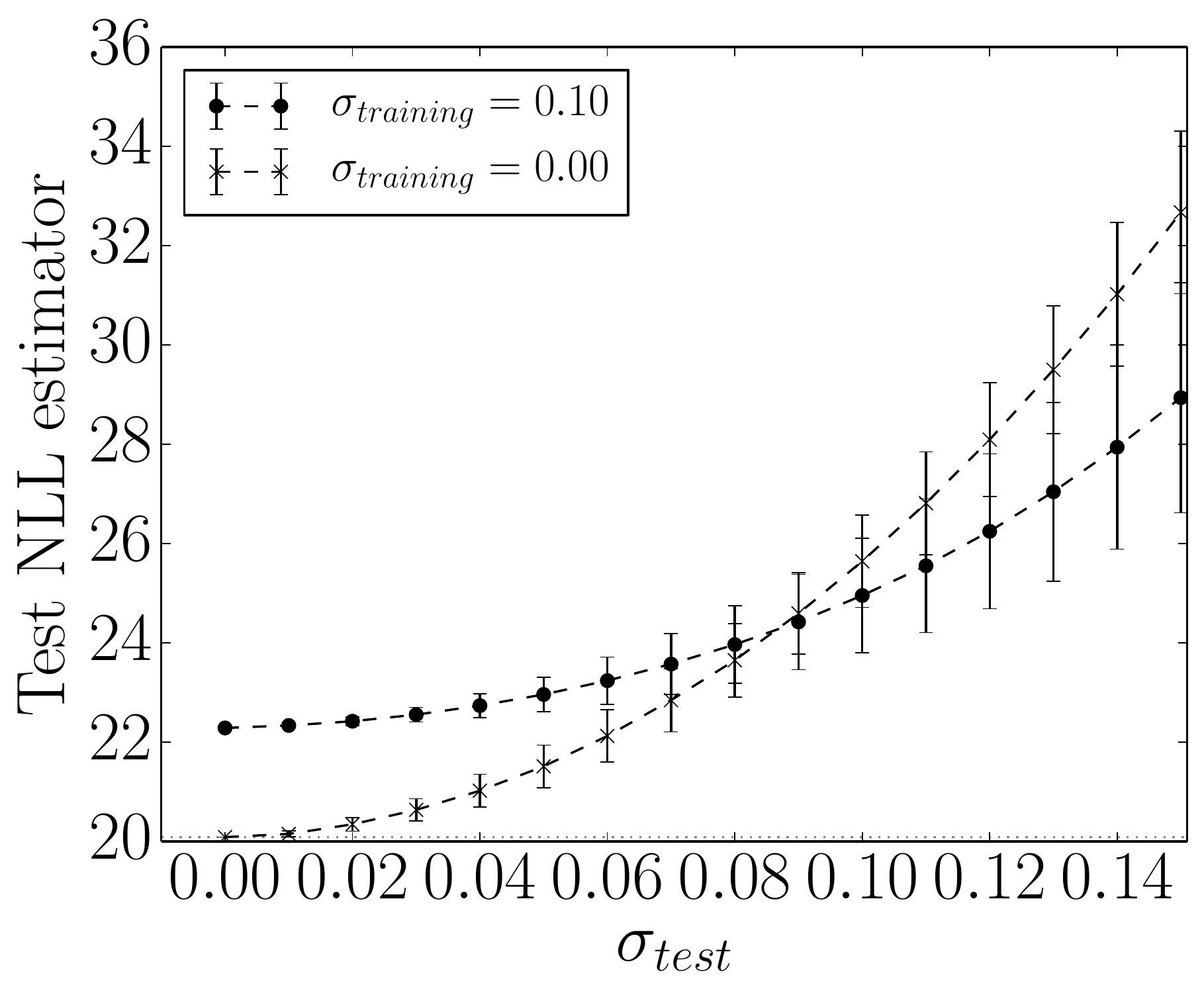} \\
	(a) & (b)
	\end{tabular}
    \caption{
a) Test negative log likelihood (NLL) estimator of a regularly-trained RBM when Gaussian
                noise is applied to parameters during sampling. For each noise
                level, NLL was computed for 5 different seeds. Noise on biases
                has practically no effect on performance compared to noise on
                weights. (Top: MNIST dataset, bottom: Connect-4 dataset)
b) Test NLL estimator of two RBMs trained with different weights
                and bias noise distributions when Gaussian noise is applied to
                parameters during sampling. For each noise level, NLL was
                computed for 5 different seeds. (Top: MNIST dataset, bottom: Connect-4 dataset)
		Qualitatively similar results were obtained for the OCR Letters dataset 
		but not displayed due to space constraints.
    }
    \label{fig:params_noise_graph}
\end{figure}

\begin{figure}[h]
    \centering
    \subfigure []{
        \label{fig:nll_graph}
        \includegraphics[width=0.47\columnwidth]{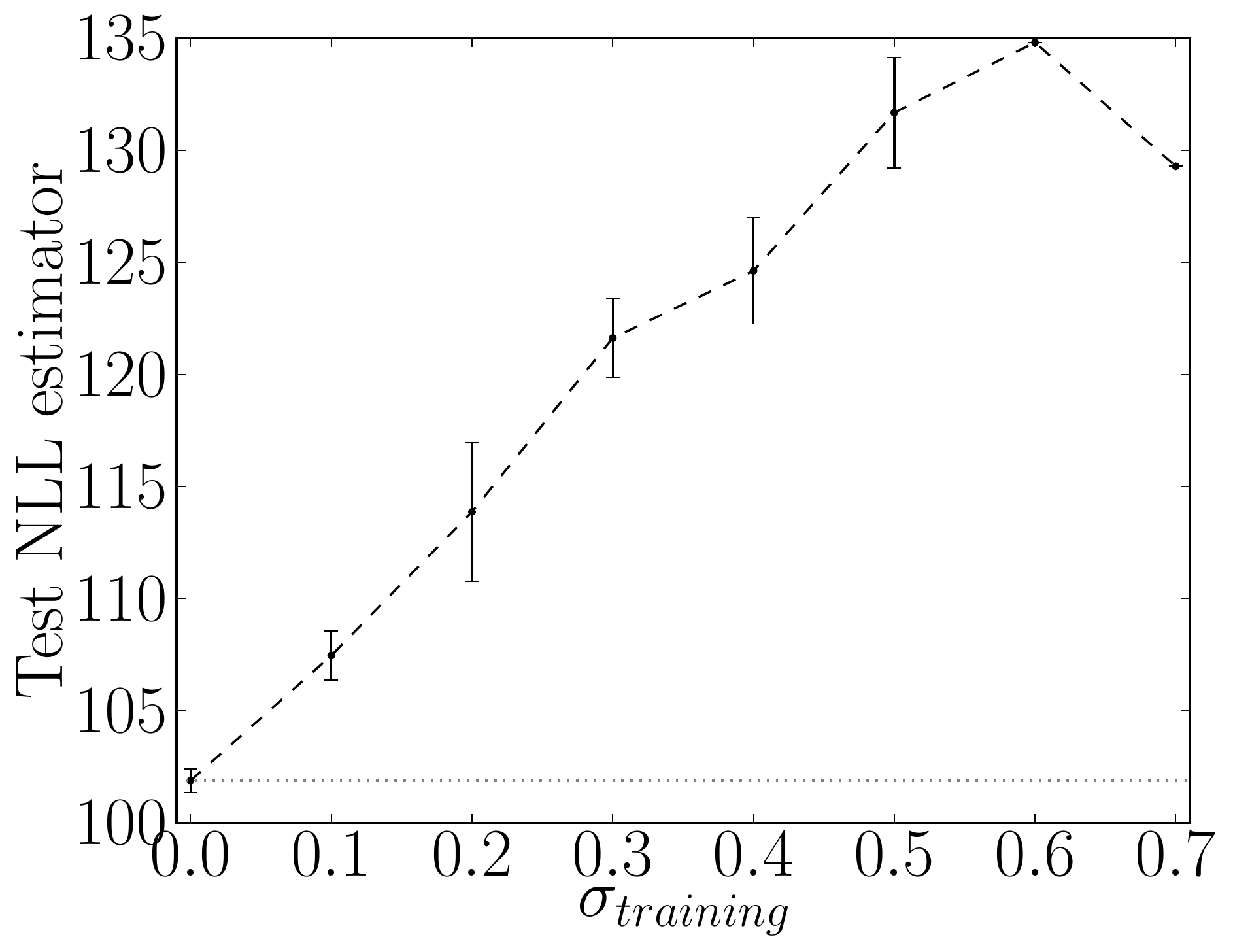}
    }
    \subfigure []{
        \label{fig:nll_amp_graph}
        \includegraphics[width=0.47\columnwidth]{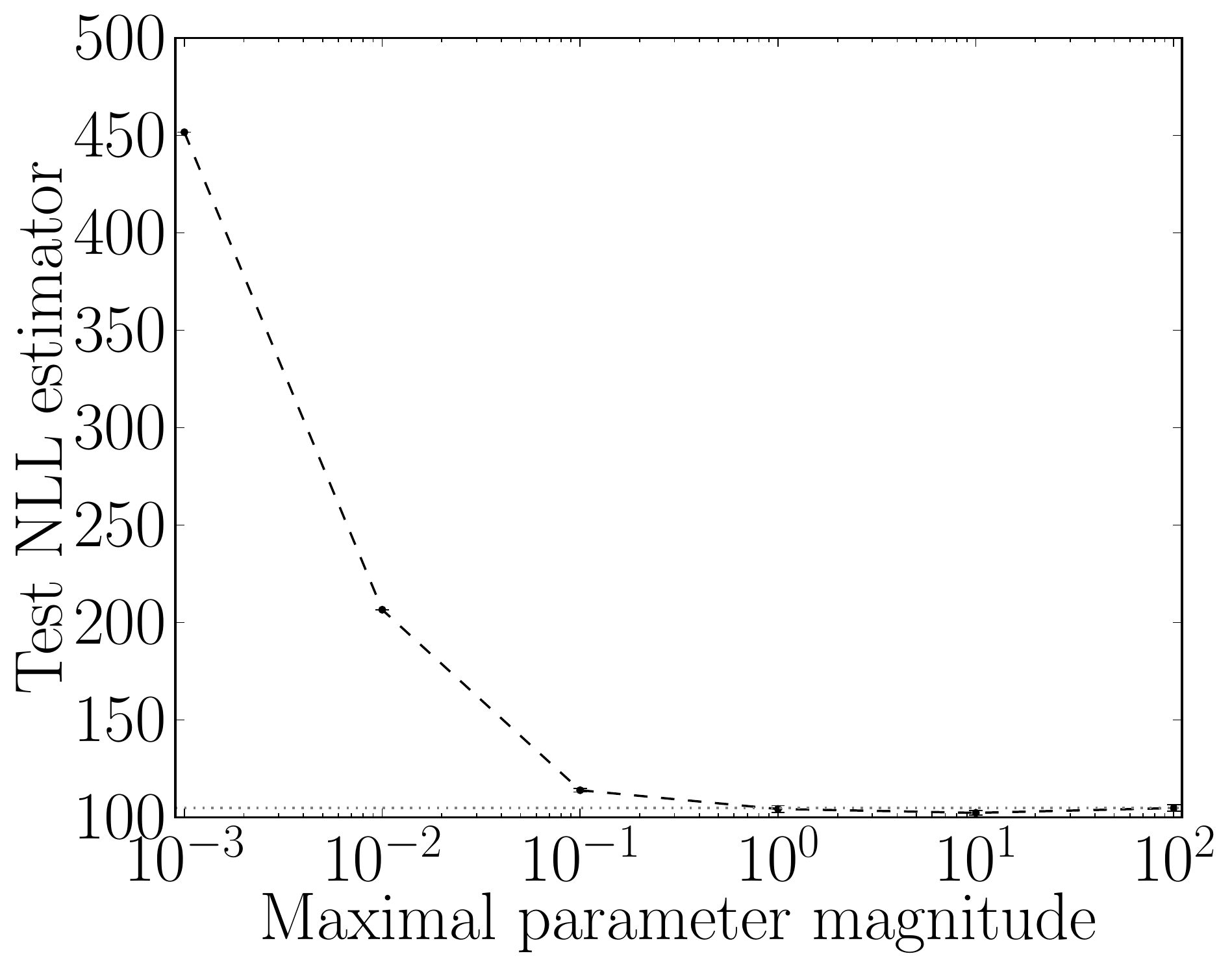}
    }
    \caption{
        a) Test NLL estimator computed by sampling with no added parameter
                noise from RBMs trained with various parameter noise levels. For
                each noise level, 5 models were trained using the same
                hyperparameters but different seeds.
        b) Test NLL estimator computed by sampling from RBMs trained with
                various magnitude constraints. For each magnitude level, 5
                models were trained using the same hyperparameters but different
                seeds.
    }
\end{figure}

\begin{figure}[h]
    \centering
        \subfigure[$\sigma_{test}=0.00$]{
            \centering
            \includegraphics[width=0.25\columnwidth]{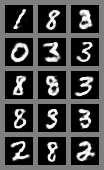}
        }
        \subfigure[$\sigma_{test}=0.10$]{
            \centering
            \includegraphics[width=0.25\columnwidth]{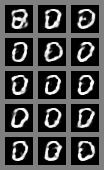}
        }
        \subfigure[$\sigma_{test}=0.15$]{
            \centering
            \includegraphics[width=0.25\columnwidth]{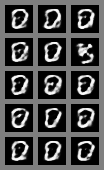}
        } \\
        \subfigure[$\sigma_{test}=0.00$]{
            \centering
            \includegraphics[width=0.25\columnwidth]{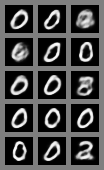}
        }
        \subfigure[$\sigma_{test}=0.10$]{
            \centering
            \includegraphics[width=0.25\columnwidth]{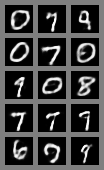}
        }
        \subfigure[$\sigma_{test}=0.15$]{
            \centering
            \includegraphics[width=0.25\columnwidth]{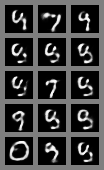}
        }
    \caption{Random samples after 100,000 Gibbs steps
             for an RBM trained without noise (top row) and for an RBM trained
             with Gaussian noise of standard deviation $\sigma=0.1$ applied to
             weights and biases (bottom row), for different levels of parameter noise.
	     }
    \label{fig:nll_samples}
\end{figure}

\begin{figure}[ht]
    \centering
        \subfigure[$\sigma_{train} = 0.10$]{
            \centering
            \includegraphics[width=0.25\columnwidth]{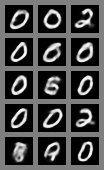}
        }
        \subfigure[$\sigma_{train} = 0.30$]{
            \centering
            \includegraphics[width=0.25\columnwidth]{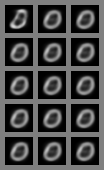}
        }
        \subfigure[$\sigma_{train} = 0.70$]{
            \centering
            \includegraphics[width=0.25\columnwidth]{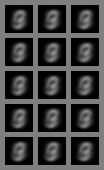}
        }
    \caption{Random samples after 100,000 Gibbs steps for
             three different parameter noise levels.
             Sampling was done without adding noise to parameters.
	     }
    \label{fig:noise_samples}
\end{figure}

\begin{figure}[ht]
    \centering
        \subfigure[Max. mag. = 100.0]{
            \centering
            \includegraphics[width=0.25\columnwidth]{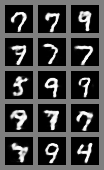}
        }
        \subfigure[Max mag. = 1.0]{
            \centering
            \includegraphics[width=0.25\columnwidth]{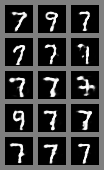}
        }
        \subfigure[Max mag. = 0.001]{
            \centering
            \includegraphics[width=0.25\columnwidth]{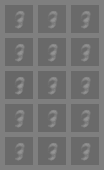}
        }
    \caption{Random samples after 100,000 Gibbs steps
             for three different parameter magnitude
             constraints.
	     }
    \label{fig:amplitude_samples}
\end{figure}

\begin{figure}[ht]
    \centering
    \subfigure []{
        \label{fig:nll_noise_amp_graph}
        \includegraphics[width=0.47\columnwidth]{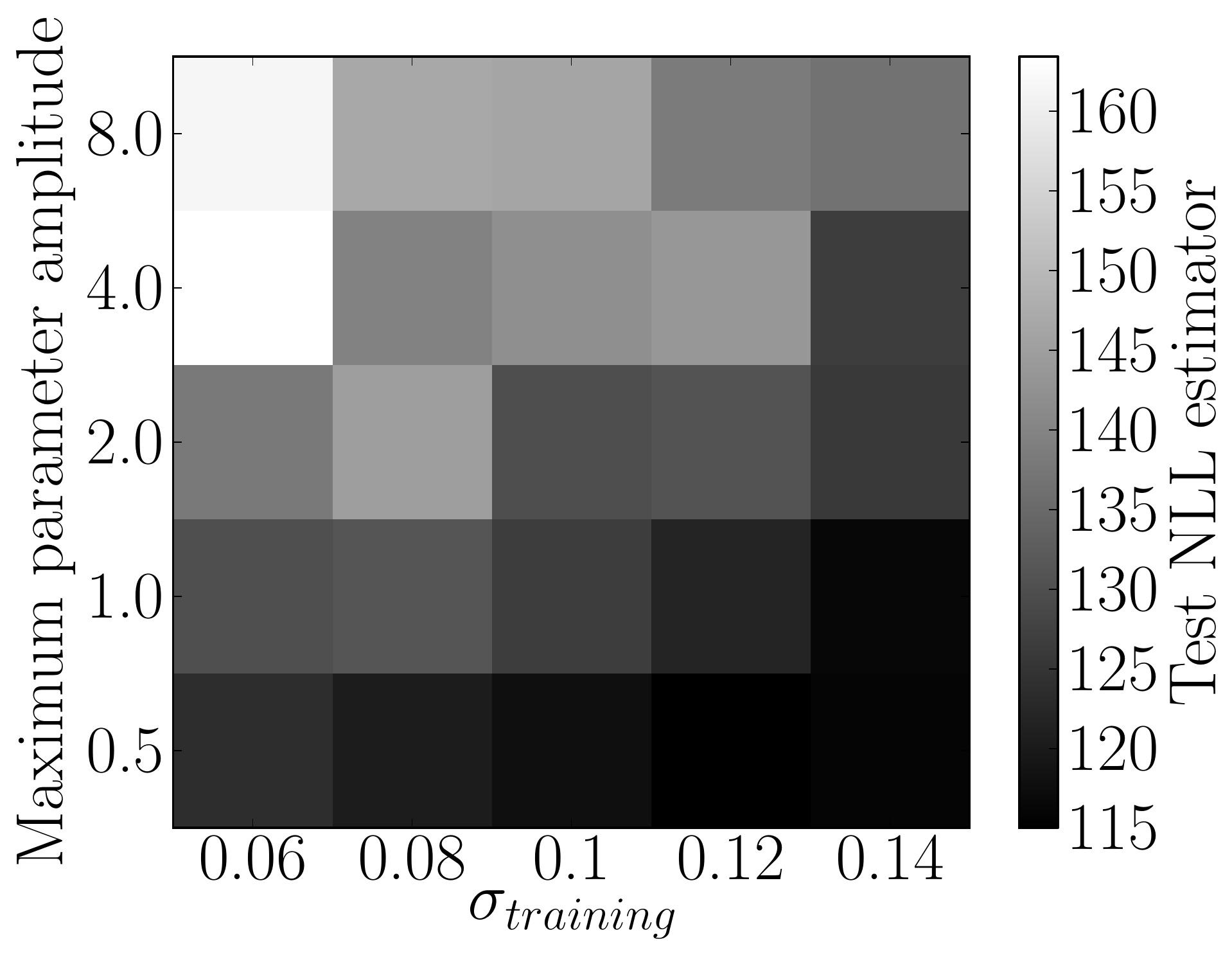}
    }
    \subfigure []{
        \label{fig:nll_arch_graph}
        \includegraphics[width=0.47\columnwidth]{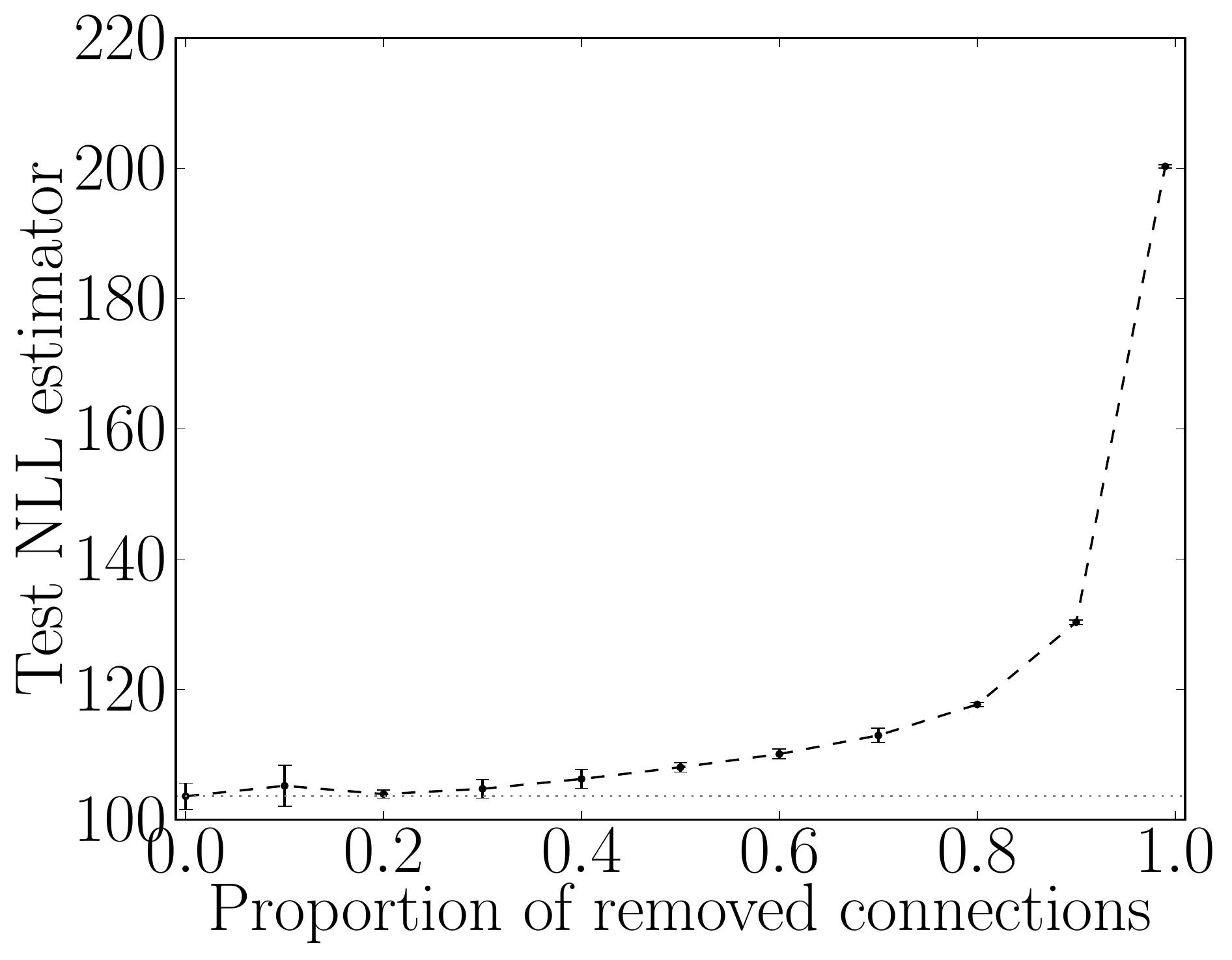}
    }
    \caption{
a) Test NLL estimator for combinations of noise and magnitude
                constraints.  In all cases, the model was evaluated using the
                same $\sigma$ as it was trained with.
b) Test NLL estimator computed by sampling from RBMs trained with
                varying amounts of removed connections (selected at random) in
                order to simulate a constrained architecture. For each
                proportion of removed connections, 5 models were trained using
                the same hyperparameters but different seeds.
    }
\end{figure}

\begin{figure}[ht]
    \centering
        \subfigure[0.10]{
            \centering
            \includegraphics[width=0.25\columnwidth]{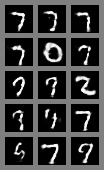}
        }
        \subfigure[0.80]{
            \centering
            \includegraphics[width=0.25\columnwidth]{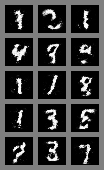}
        }
        \subfigure[0.99]{
            \centering
            \includegraphics[width=0.25\columnwidth]{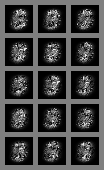}
        }
    \caption{Random samples after 100,000 Gibbs steps
             for three different proportions of removed
             connections.
	     }
    \label{fig:architecture_samples}
\end{figure}

\begin{figure}[ht]
    \centering
        \subfigure[]{
            \centering
            \includegraphics[width=0.25\columnwidth]{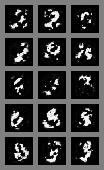}
        }
        \subfigure[]{
            \centering
            \includegraphics[width=0.25\columnwidth]{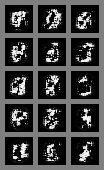}
        }
    \caption{Random samples after 100,000 Gibbs steps
             for two RBMs trained with a chimera connectivity pattern, using
             a) \emph{pixel blocks} and b) \emph{extended pixel blocks}
             pixel-to-units mappings.
	     }
    \label{fig:chimera_samples}
\end{figure}

%------------------------------------------------------------------------------
%                               MAIN SECTION
%------------------------------------------------------------------------------
\section{Methodological notes}

All models were trained using PCD-15.
We used standard train / test split for both the MNIST~\cite{LeCun+98} 
and Connect-4 and OCR Letters~\cite{Larochelle+al-2010} datasets.
For all experiments involving training on the simulated physical computer,
we used the simulator to draw samples for the negative phase of PCD, but used exact mean
field for the positive phase.
Training examples were binarized every time they were presented by
sampling from a Bernoulli distribution, such that the grayscale value in [0, 1] in the
original image gives the probability of that pixel being a 1 in the binary image.
Unless explicitly stated, all models were
trained using the same hyperparameters.

{\em Negative log-likelihood} (NLL) of all models is approximated using annealed
importance sampling (AIS) ~\cite{Salakhutdinov+Murray-2008}.
When noise is added to parameters, the expected AIS is computed by Monte Carlo,
with test examples binarized by following the same method as with training
examples.

Although the constraints we apply are dictated by the D-Wave Two system, we are
simulating a low-precision RBM, which means that constraints are enforced on
parameters directly, without converting them to the Ising parametrization first.

All images of samples are displaying the {\em expected value} of the visible units
given binary samples of the hidden units.

%------------------------------------------------------------------------------
\section{Simulating noisy parameters}
\label{noise}

Consider the case where we have a trained RBM (trained by any succesful means;
in these experiments we obtained ours by traditional
training on a digital computer), and we would like to draw samples
from it using physical computation. In this case, we know that the model
parameters represent the desired distribution well. However, when loaded into
the physical computer, the parameters may not be preserved exactly. We simulate
this by adding Gaussian noise to the parameters.

See Fig. \ref{fig:params_noise_graph} for a summary of the experimental results in
this case.  We find that noise on biases has a negligible effect on NLL compared
to noise on weights. This could be explained by the fact there are simply more
weight parameters than bias parameters contributing to the energy function.  In
that case, variance of the energy function would be dominated by variance on
weights.  From these tests, we can observe two things: 1) Adding noise to the
model parameters quickly degrades its performance, and 2) Noise on the biases is
less harmful than noise on the weights.

Of course, these parameters were trained to work well in the absence
of noise.  It is possible to learn different parameters, that are chosen to
diminish the effect of noise. In order to do this, we trained an RBM using
the simulated physical computer to draw the negative phase samples during
training. The negative phase repels the model parameters from regions that
produced poor samples. Using noise on the parameters while generating the
negative phase samples increases the range of the repulsion--not only must the
parameters not generate bad samples, noisy versions of the parameters must not
do so either.

We compared how RBM performance evolves as we increase parameter noise during
sampling with that of the RBM trained without noisy parameters (Fig.
\ref{fig:params_noise_graph}).
We used the same noise distribution for both weights and biases.

We find that training with noisy parameters helps reducing the degrading effect
of sampling with noisy parameters. For instance, by training with $\sigma =
0.10$ on parameters and sampling with the same $\sigma$, we were able to reduce
NLL estimator increase by 21.9\% in average when compared to training without
noise.  Furthermore, the benefits of training with noisy parameters before
sampling with noisy parameters extends to noise levels greater than used during
training.

The effect of training with noisy parameters is also qualitatively visible when
looking at samples from the model (Fig. \ref{fig:nll_samples}).  We
observe that adding noise to parameters during sampling increases visual noise
in samples, and also makes samples collapse to major modes. By training with
noisy parameters, we are able to soften these effects, even when sampling with
parameter noise greater than training noise.

As for how much parameter noise an RBM can support during training, we trained
RBMs using various noise levels on weights and biases and computed their test
NLL estimator when sampling with no added noise (Fig. \ref{fig:nll_graph}).
A noise level of $\sigma = 0.1$ is the biggest noise we could add before the
RBM's performance noticeably started to degrade. This means noisy parameters
negatively affects learning for all but the smallest noise values.

%------------------------------------------------------------------------------
\section{Simulating limited parameter range}
\label{range}
We now turn our attention to the parameter range constraint. We trained RBMs by
forcing their parameter magnitude to stay below a certain threshold value and
observed the effect of that value on test NLL (Fig. \ref{fig:nll_amp_graph}).
Whenever parameter updates would bring a parameter outside of that range, it was
clipped to the threshold value.

We find that a magnitude constraint higher than or equal to 1.0 has little to no
effect on performance, but that forcing parameter's magnitude to be smaller than
that quickly degrades performance.

%------------------------------------------------------------------------------
\section{Combined simulation of noise and limited parameter range}
\label{noise_range}
We combined noise and magnitude constraints together to see how they interact.
We explored constraint space around reasonable noise and magnitude values and
looked at how they affect NLL (Fig. \ref{fig:nll_noise_amp_graph}). The two
constraints appear to work well together. In fact, a model with higher noise and
small parameter values performs nearly as well as a standard RBM. We think that
the constraint on parameter values may actually be helpful, because they force
the RBM to find good weight vector directions that generalize well, rather than
just scaling up its weights to overpower the noise. As always, one should be
careful about generalizing these conclusions to values outside the ranges evaluated
in these experiments.

%------------------------------------------------------------------------------
\section{Simulating limited connectivity}
\label{connectivity}
We trained RBMs by forcing a random subset of weights to be zero and observed
how it affected test NLL (Fig. \ref{fig:nll_arch_graph}).
It turns out the RBM can cope with a reasonable amount of removed connections:
even when half the weights are forced to be zero, test NLL only increases by
about 4.3\%. However, physical implementations will likely have sparse
connectivity; for instance, the connectivity pattern of a D-Wave machine (Fig.
\ref{fig:pixels_to_chimera}) applied to an RBM with 784 visible units and 784
hidden units is so that over 99\% of its connections are removed. In the
aforementioned experiment, 99\% removed connections results in a
disappointing $200.3 \pm 0.2$ test NLL. When looking at samples (Fig. \ref{fig:architecture_samples}), we observe that
the RBM's representative power decreases as we force more weights to be zero,
until samples no longer resemble digits.

Fortunately, physical implementations of an RBM will most likely have some
kind of structure to their connectivity pattern, so the results we get
by forcing a random subset of the weights to be zero are somewhat pessimistic.

When we train an RBM with 784 visible units and 784 hidden units with
chimera connectivity pattern, results are much better. There are many ways to
map pixels of an image to visible units of the model; we tried two that seemed
the most logical (Fig. \ref{fig:pixels_to_chimera}). The \emph{pixel blocks}
mapping lead to a test NLL of 138.2, while the \emph{extended pixel blocks}
mapping lead to a test NLL of 160.9. When we look at samples from both RBMs
(Fig.  \ref{fig:chimera_samples}), we see that digit structure is much better
preserved than when we randomly force the same proportion of weights to be zero,
although samples still barely look like digits.  In all cases, the limited
architecture seems to be the most damaging constraint studied in this paper.

%------------------------------------------------------------------------------
%                                CONCLUSION
%------------------------------------------------------------------------------
\section{Conclusion}
\label{conclusion}

In this paper, we have performed a series of simulation experiments to determine
the feasibility of implementing an RBM using physical computation. We have
evaluated the impact of three barriers to the success of physical computation:
noise on the model parameters, limited range on the model parameters, and
limited topology of the model.

We have found that noise on the parameters noticeably degrades performance,
though this can be mitigated by training using the same sampler in the negative
phase as will be used to draw samples at test time. We have found that the
limits on the range of the parameters do not significantly impair the
performance of the RBM. Finally, and most importantly, we have found that
restrictions on the topology of the model can impair the model's performance
more than any of the other limitations we consider.
While structured sparsity like in the D-Wave Two system's chimera topology does
perform well for the number of connections it has, the overall number of
connections is still low enough to cause many difficulties.

Note however that experiments on noisy weights were performed on fully-connected
RBMs.  If, as suggested when discussing Fig.  \ref{fig:params_noise_graph}, the effect
of noisy parameters is dominated by noise on weights because there are more
weights than biases, then a constrained architecture might mitigate the effect
of noisy weights simply by reducing their number. This needs to be verified in
future experiments.

This suggests that quantum hardware designers should concentrate their efforts
on reducing noise on weights and on increasing the number of connections between
elements in the quantum computers, and quantum machine learning researchers
should focus their efforts on designing approaches that can cope with noisy
weights and restricted topology.

%------------------------------------------------------------------------------
%                                REFERENCES
%------------------------------------------------------------------------------
\pagebreak
\bibliography{strings,strings-shorter,ml,aigaion}
\bibliographystyle{aaai}

\end{document}